\documentclass{easychair}

\usepackage[T1]{fontenc}
\usepackage{underscore} 
\usepackage{hyperref}
\usepackage{xspace}
\usepackage{enumitem}
\usepackage{float}
\usepackage{parskip}
\usepackage{graphicx}
\graphicspath{{figures}}
\usepackage{booktabs}
\usepackage{multirow}
\usepackage{placeins}
\usepackage{calc}

\usepackage{tikz}
\usetikzlibrary{shapes.geometric, arrows.meta, positioning, calc}

\usepackage{pgfplots} 
\usepackage{pgfplotstable}
\pgfplotsset{width=15.5cm, height=6.5cm, compat=1.18}  

\usepackage[frozencache]{minted}
\setminted[prolog]{
  fontsize=\small,
  baselinestretch=0.9,
  mathescape=true,
  linenos=true
}

\setminted[text]{
  fontsize=\small,
  baselinestretch=0.9,
  mathescape=true,
  linenos=true
}

\newcommand{\pl}[1]{\text{\mintinline{prolog}{#1}}}
\newcommand{\txt}[1]{\text{\mintinline{text}{#1}}}

\usepackage{tabularx}
\newcolumntype{L}{>{$}l<{$}}%
\newcolumntype{C}{>{$}c<{$}}%
\newcolumntype{R}{>{$}r<{$}}%
\newcolumntype{M}{@{}p{\mathindent}@{}}%
\newcolumntype{t}{>{\mbox\bgroup}l<{\egroup}}%
\newcolumntype{e}{r@{\;}l}%
\newcolumntype{E}{>{$}r<{$}@{$\;$}>{$}l<{$}}%

\theoremstyle{definition}

\usepackage{tcolorbox} 
\tcbuselibrary{skins,theorems} 
\newtcbtheorem{algorithm}{Algorithm}{pseudo/ruled}{alg}

\usepackage{pseudo}
\usepackage{tabto}
\TabPositions{1.5cm}
\pseudoset{
ct-left={{~\small//~}},
ct-right=\relax,
indent-length=1em,
cnfont=\sffamily,
line-height=1.1,
label={\small\arabic*},
kw,
}

\pseudodefinestyle{fullwidth}{
    begin-tabular =
\tabularx{\linewidth}[t]{@{} r
>{\pseudosetup}
X >{\leavevmode\small\color{gray}} p{0.4\linewidth}
@{}},
    end-tabular = \endtabularx,
    setup-append = \RestorePseudoEq
}

\newcommand{\addSpace}[3]{\rule[-\dp#1 - #2]{0pt}{\dp#1 + #3 + \ht#1}}

\definecolor{paleyellow}{HTML}{FFE3C0}
\definecolor{dullblue}{RGB}{100,100,200}
\long\def\comment[#1]#2{\par\colorbox{paleyellow}{\llap{#1:\quad}%
    \parbox[t]{\textwidth}{\setlength{\parskip}{1ex plus 0.2ex minus 0.2ex}#2}}}

\def\void{}
\newcommand{\mcomment}[2][\void]{\marginpar{\raggedright\footnotesize\ifx\void#1\else
      {\fontsize{6pt}{6pt}\selectfont #1:} \fi#2}}

\pgfplotstableread{
Y    PosErr  NegErr
0	0	0
0	0	0
13.3	6.7	13.3
0	0	0
13.1	2.1	4
1.9	3.7	1.9
}\datatableXZero

\pgfplotstableread{
Y    PosErr  NegErr
3.8	1.1	0.7
14.9	11.4	7.5
15	10	15
0	0	0
9.1	2.3	4.1
8.1	10.6	8.1
}\datatableXOne

\pgfplotstableread{
Y    PosErr  NegErr
6.1	6.4	3.7
1.8	3.5	1.8
0	0	0
17.9	10.7	17.9
10.1	2.4	1.4
16.1	6.1	6.7
}\datatableXTwo

\pgfplotstableread{
Y    PosErr  NegErr
10.7	4.9	4
6.7	0.7	1.4
0	0	0
0	0	0
8.4	5.3	3.4
3.2	3.1	3.2
}\datatableXThree

\pgfplotstableread{
Y	PosErr	NegErr
35.2	5.4	3.5
40.7	3.8	3.8
78.3	1.7	3.3
63.1	11.9	20.2
43.2	1.8	1.9
51	2.3	4.1
}\datatableXFour

\pgfplotstableread{
Y	PosErr	NegErr
94.9	5.1	7.1
79.4	9.5	9
15	10	15
27.4	1.2	2.4
84.6	0.4	0.5
85.3	8.1	10.3
}\datatableXFive

\usepackage{url}
\usepackage{natbib}
\setcitestyle{numbers,square,comma} 

\definecolor{DodgerUniformBlue}{rgb}{0.0,0.353,0.612}

\title{Automated Theorem Provers Help Improve Large Language Model
  Reasoning\footnote{This paper has also been published in: N. Bjørner, M. Heule and
    A. Voronkov (eds.), LPAR 2024 (EPiC Series in Computing, vol. 100), pp. 51–69}}
\titlerunning{Automated Theorem Provers Help Improve LLM Reasoning}

\author{Lachlan McGinness\inst{1} \and Peter Baumgartner\inst{2}}
\authorrunning{McGinness and Baumgartner }
\institute{}

\institute{
  School of Computer Science, Australian National University and Data61, CSIRO \\
  \email{lachlan.mcginness@anu.edu.au}
  \and
  Data61, CSIRO 
  and
  School of Computer Science, Australian National University \email{peter.baumgartner@data61.csiro.au}
}

\begin{document}

\maketitle

\begin{abstract}
In this paper we demonstrate how logic programming systems and Automated first-order logic
Theorem Provers (ATPs) can improve the accuracy of Large Language Models (LLMs) for
logical reasoning tasks where the baseline performance is given by direct LLM
solutions. We first evaluate LLM reasoning on steamroller problems using the PRONTOQA
benchmark. We show how accuracy can be improved with a neuro-symbolic architecture where
the LLM acts solely as a front-end for translating a given problem into a formal logic
language and an automated reasoning engine is called for solving it.
However, this
approach critically hinges on the correctness of the LLM translation.
To assess this translation correctness, we secondly define a framework of syntactic and semantic error
categories. We implemented the framework and used it to identify errors that LLMs 
make in the benchmark domain.
Based on these findings, we thirdly extended our method with capabilities for automatically
correcting syntactic and semantic errors. For semantic error correction we integrate
first-order logic ATPs, which is our main and novel contribution.
We demonstrate that this approach reduces semantic errors significantly
and further increases the accurracy of LLM logical reasoning.
\end{abstract}

\section{Introduction, Background and Related Work}
The release of models like GPT \cite{Brown2020Language} and Gemini \cite{Google2023Gemini} through platforms like ChatGPT and Bard have transformed Large Language Models (LLMs) into general-purpose tools that can be used by everyone. 
Although designed for next token prediction, LLMs have been shown to have emergent abilities and are able to perform a wide variety of tasks without task-specific training data \cite{Brown2020Language, Polu2020Generative, Srivastava2023Beyond, Wei2022Emergent, Wei2022Chain}. 

Unfortunately, LLMs also frequently return wrong results, such as fictitious claims  (``hallucinations'') or conclusions that defy common sense or (naive qualitative) physics \cite{Liu2023Minds, Mailon2023Augmented, Tafjord2019Quarel}. Such shortcoming may or may not be obvious but in any case impact trustworthiness. A recent famous example was a lawyer who submitted a legal brief generated by ChatGPT which contained many errors and false references \cite{dahlHallucinatingLawLegal2024, dahlLargeLegalFictions2024}. Asking the LLM for an explanation might help, but the explanation might contain errors again and does not necessarily reflect the process used to obtain its answer. Equipping and checking LLMs with trustworthy (logical) reasoning
remains to be a current major problem \cite{Press2023Measuring, Qiao2023Reasoning}.

A general approach to address this problem equips LLMs with external functionality~\cite{Gao2023PAL, Kassner2023Language, Liu2023Minds, Poesia2023Certified, Press2023Measuring}. These equipped models are referred to as Augmented Language Models (ALMs). The general problem of combining neural networks with symbolic reasoners has
attracted a lot of attention recently (a popular umbrella term is ``neuro-symbolic
computation''). An impressive example is the work by Trinh et\ al. \cite{AlphaGeometry} which
demonstrates that a neuro-symbolic architecture that can solve International Mathematics Olympiad geometry
questions at an expert level.

Other proposed combination schemes range from end-to-end differentiable architectures
with tightly integrated training regimes \cite{DeSmet2023Neural, Manhaeve2018DeepProbLog, Manhaeve2021Neural, Winters2022DeepStochLog} to more loosely coupled systems where pre-trained models are linked with a reasoner through a formal language interface \cite{Rajasekharan2023Reliable, Tafjord2019Quarel}. In this paper we consider combinations of the latter kind. Pre-trained LLMs are used as black boxes tasked with translating problems that require logical reasoning into a formal logic, so that an Automated Reasoning (AR) system can be applied. We first show how accuracy can be improved with such a neuro-symbolic architecture. 

This approach naturally provides excellent explainability and trustworthiness on the AR side. Therefore
the correctness of the overall system critically hinges on the correctness of the LLM translation.
However, engineering prompts with high correctness requires many test cases and iterations. As a result, manual inspection of test case results quickly becomes
unfeasible. Knowing the types of errors that the LLM makes has the capacity to inform prompt engineers allowing optimal performance to be reached more quickly.

In order to assess the correctness of the LLM translation from natural language into logic
programs, we need a reliable ground-truth logic representation for the natural language problem.
To make this possible, we follow current approaches and work in a controlled setting. We
chose popular ``steamroller'' problems, which are readily available in useful variants and
can be auto-generated in any number \cite{Saparov2023Language}. We wrote a standard Definite Clause Grammar (DCG) parser for the required
subset of English and that outputs First Order Logic (FOL) formulas, our ground truth formulas.
This puts us in a position to compare the two formal logic representations; the first from the LLM and the second from the DCG. We do that in a purely semantic way using \pr{SEDAC} (\emph{Semantic Error Detection And Correction}); an algorithm that calls an Automated Theorem Prover (ATP) that is capable of deciding entailments in the considered fragment for two given formalizations.

We are interested in analyzing the correctness beyond a binary true/false status.
In case of incorrectness, we make certain modifications
to the given formulas and check again for entailment. Depending on the result, this allows us to
conclude certain error classes and carry out automatic corrections.

We can illustrate our approach with a metaphor from text processing. Virtually every
natural language text editor includes a spell-checker for (a) fixing spelling mistakes and
(b) grammatical errors. More recently, (c) semantic analysis for, e.g., finding the right
words for a given writing style have been added.  Roughly speaking, our error categories
correspond to these three levels. We have syntax errors (a), shallow semantic errors (b),
and deep semantic errors (c). Like in text processing, they come at different levels of
automatic detectability, fixability and the need to validate the proposed fix with the
user or environment.

As far as we know, ours is the first approach of its kind.
We describe it and report on practical experiments. 
The approach and the result statistics are valuable for at
least three reasons: 
(1) they provide insights into expected problem areas of ALMs that are generalizable, (2) they can give a human in the loop insights to create targeted improvements to LLM prompts, and (3) they offer the ability to `auto-correct' some types of semantic errors made by LLMs when calling tools leading to improved performance.

\paragraph{Related work.}
The reasoning capability of LLMs is an active area of research, we refer to Huang et\ al.\ (2023) for a general overview of this field \cite{Huang2023Towards}. The majority of this work focuses on enhancing LLM reasoning capabilities with fine-tuning and prompt engineering but without the use of external reasoning tools. Key methods include Chain of Thought (CoT) reasoning \cite{Wei2022Chain}, zero-shot CoT reasoning \cite{Kojima2022Large}, Selection-Inference \cite{Cresswell2023Selection} and backward chain reasoning \cite{Kazemi2023LAMBADA}. 

Although there are many works which measure the performance of LLMs on logical reasoning
benchmarks \cite{Kojima2022Large, Mailon2023Augmented, Petroni2019Language}, very little
work has been done to classify the types of errors they make. Xu et\
al.~\cite{Xu2023Large} focuses on the emergent reasoning capabilities of LLMs (a fully
sub-symbolic approach) and proposes two classes; evidence selection errors and reasoning
process errors. These categories are not appropriate for neuro-symbolic approaches such as
ALMs which allow models to make use of external tools for logical reasoning
\cite{Mailon2023Augmented}. In these approaches, reasoning process errors are not
relevant, instead the LLM is required to select the correct evidence and successfully
translate it into instructions to be parsed by an external tool. Therefore for this domain
we propose different error classes: syntactic errors and semantic errors, see Section~\ref{sec:errorcat}. 

\section{Our Method}
\label{sec:methods}

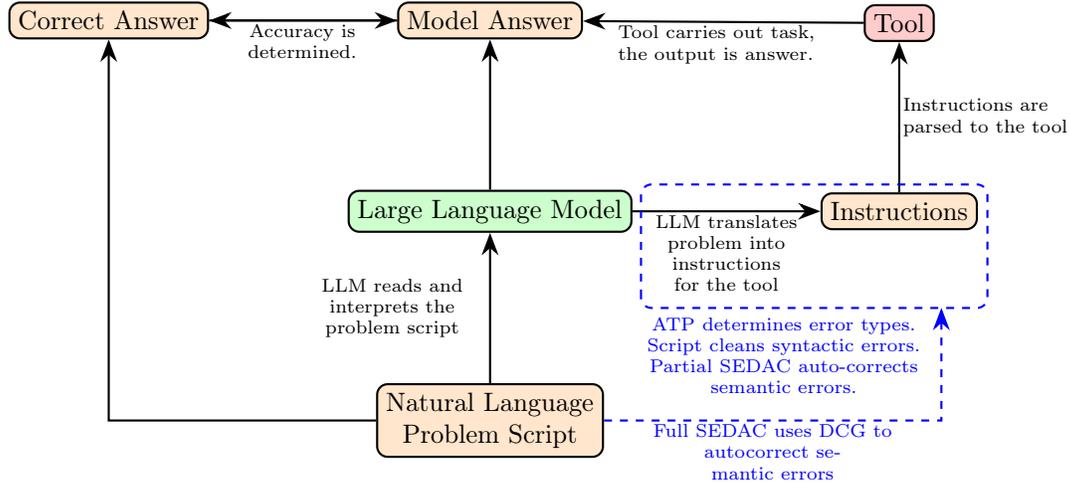
\begin{figure}[h!]
    \caption{A diagram of the general structure of Large Language Model tool use. In order to successfully use a tool an LLM must successfully generate instructions for that tool that are free of both syntax errors and semantic errors. Our contributions to improving this process including auto-correcting and error type classification are shown blue.}
    \centering
    \begin{tikzpicture}[node distance=2cm and 2.5cm,
    box/.style={rounded corners, draw, align=center, thick, fill=#1!20},
    arrow/.style={-{Stealth[length=3mm]}, thick},
    arrowlabel/.style={font=\scriptsize, fill=none, inner sep=1pt}]

    \node[box=orange] (problem) {Natural Language\\ Problem Script};
    \node[box=green, above=of problem] (llm) {Large Language Model};
    \node[box=orange, right=of llm] (instructions) {Instructions};
    \node[box=red, above=of instructions] (tool) {Tool};
    \node[box=orange, above=of llm] (output) {Model Answer};
    \node[box=orange, left=of output] (truth) {Correct Answer};

    \draw[arrow] (problem) -- (llm) node[arrowlabel, pos=0.5, left, align=center, text width=2.5cm] {LLM reads and interprets the problem script};
    \draw[arrow] (llm) -- (output) node[arrowlabel, pos=1.5, left] {};
    \draw[arrow] (instructions) -- (tool) node[arrowlabel, pos=0.5, right, text width=2.5cm] {Instructions are parsed to the tool};
    \draw[arrow] (tool) -- (output) node[arrowlabel, pos=0.5, below, text width=2.8cm] {Tool carries out task,\\ the output is answer.};
    \draw[arrow] (problem) -| (truth) node[arrowlabel, pos=0.5, above] {};
    \draw[arrow] (truth) -- (output) node[arrowlabel, pos=0.5, above] {};
    \draw[arrow] (output) -- (truth) node[arrowlabel, pos=0.5, above] {};
    \draw[arrow] (llm) -- (instructions) node[arrowlabel, pos=0.5, below, align=center, text width=2.3cm] {LLM translates problem into \\instructions for the tool}; 
    \draw[arrow, blue, dashed] (problem) -| (6,1.5) node[arrowlabel, pos=0.25, below, text=blue, align=center, text width=3.3cm] {Full SEDAC uses DCG to \\autocorrect semantic errors}; 
    
    \draw[blue, dashed, thick, rounded corners] ($(llm.south west)+(3.9,-1.)$) rectangle ($(tool.north east)+(0.7,-2.4)$);
    \node[align=center, font=\scriptsize, blue, anchor=north] at ($(llm.south west)+(5.8,-1.0)$) (caption) {ATP determines error types. \\ Script cleans syntactic errors. \\ Partial SEDAC auto-corrects \\semantic errors.};
    
    \node[align=center, font=\scriptsize, black, anchor=north] at ($(llm.south west)+(-0.60,2.9)$) (caption) {Accuracy is\\ determined.};
    
\end{tikzpicture}
    \label{fig:LLMtoolpipeline}
\end{figure}

Natural Language Processing is a fast moving area with multiple new LLMs being released
each year. This work focuses on only three of the best performing models at the time
of the experiment; GPT3.5 \cite{Brown2020Language}, GPT4 \cite{Openai2023GPT4} and Gemini-Pro
\cite{Google2023Gemini}. This study investigates the logical reasoning skills of these
models and how they could be augmented through the use of automated reasoning systems. Figure~\ref{fig:LLMtoolpipeline} provides an overview of the general architecture that we explore in our experiments.

To test these models we chose PRONTOQA \cite{Saparov2023Language}, a logical reasoning dataset, because it has different settings (`ontologies', `hops' and `distractors') which can be changed to adjust the difficulty of the problems. PRONTOQA provides the Natural Language Problem Script for our specific experiments. The code for PRONTOQA questions is published but not the questions themselves, which helps prevent contamination of LLMs (reduces the likelihood that they will have seen the exact questions and answers in their training data). We generated one hundred examples of the most difficult problems (`false ontology' with `relevant distractors') for one hop, two hops and three hops as our evaluation benchmark.  

We implemented several experimental conditions for each LLM. In the baseline condition the model was given a question from the benchmark and needed to produce a `True' or `False' answer based on the text provided. This corresponds to the arrow pointing from the Large Language Model to the Model Answer in Figure \ref{fig:LLMtoolpipeline}. For the zero-shot condition, we provide the LLM with instructions explaining how to write a Logic Program (LP) in Prolog syntax and ask it to convert a natural language problem
into such a logic program. The LP \textit{is} the instructions shown in Figure \ref{fig:LLMtoolpipeline}.

We chose logic programs as the interchange language because their syntax is already known by the LLMs, they are easy to ``teach'' to a LLM in a prompt and their simple if-then structure is sufficient for our purpose.
For computing a `True' or `False' result we used our Fusemate LP system \cite{baumgartner-bottom-up-2023-3}.\footnote{The PRONTOQA problems are designed in such a way that both an open world or closed world semantics based reasoner can be used with the same result.}

For our specific case, Fusemate is the Tool illustrated in Figure \ref{fig:LLMtoolpipeline} that produces the Model Answer. The arrow pointing from Large Language Model to the Instructions is the pipeline that we are evaluating.

For the one-shot condition, we provide the LLM with instructions for how to write a
logic program, an example natural language problem, the corresponding logic program and a new natural language problem. Once
again the resulting logic program is sent to Fusemate to compute a `True' or `False'  answer. We repeated this process for each problem in the benchmark and for each of the three LLMs.

We generated error logs from each trial which contain each problem, the corresponding
model answer, the correct answer and the logic programs generated by the models
for the zero-shot and one-shot conditions. These experiments and their result statistics
revealed several weaknesses with these approaches in terms of the error framework
introduced above.

\subsection{Error Categorisation}
\label{sec:errorcat}

Few systems for error categorisation currently exist in the literature \cite{Xu2023Large} and these are not appropriate for categorising errors when Augmented Language Models (ALMs) call upon tools. Therefore we propose a new error categorisation which has two broad classes; syntactic errors and semantic errors.

A Syntactic Error is defined as an error in the LLM's instructions which prevents the tool from parsing. There are a number of different sub-categories of syntactic error which can contribute to this including:
\begin{itemize}
    \item \textbf{Symbol Errors} - The LLM instructions contain incorrect symbols. As an example consider a logic program which contained ``-?''  instead of ``?-'' for a query. This would prevent the script from running and so the instructions can no longer be parsed. 
    \item \textbf{Natural Language Errors} - The model includes natural language in addition to or instead of machine interpretable instructions.
    \item \textbf{Communication Errors} - A specific form of Natural Language Error where the model incorporates markers like ```''' or <<>> to separate natural language from tool instructions. A human can very easily interpret which parts are meant to be included in the instructions. This type of error can be cleaned very easily.  
    \item \textbf{Knowledge Errors} - This is a form of evidence selection error \cite{Xu2023Large}. Rather than translating the problem directly, the LLM tries to incorporate some of its own pre-existing knowledge into the instructions. An example is when the model replaces `even number' with `integer divisible by 2'.
    \item \textbf{Other Syntax Errors} - Any other syntactic error that prevents the model from parsing which does not fall into the categories above. This is depended on the tool being used. 
\end{itemize}

A Semantic Error is an error in which the instructions are able to be parsed by
the tool but give an incorrect output.
The exact types of semantic errors are tool dependent.
However we recommend breaking these into two sub-categories that will likely be helpful
to developers:

\begin{itemize}
    \item \textbf{Shallow Semantic Errors} - Errors where the semantic meaning can confidently be recovered (auto-corrected) without viewing the original natural language script. We suggest that these could be referred to as auto-correctable errors.
    \item \textbf{Deep Semantic Errors} - Errors where the semantic meaning cannot be recovered without viewing the original natural language script. We suggest that these could be referred to as non-auto-correctable errors.
\end{itemize}

Establishing a system of well defined error categories provides a common language and allows focus on specific common errors for the NLP community to address. This error classification is also important for developers to identify the best technique to improve the performance of LLMs. For example, if a  developer 
discovers a large number of syntactic errors then they know to focus on techniques that can reduce these:  one or few-shot prompts, fine-tuning the model with a focus on the tool's grammar or writing a script that will correct syntax on LLM instructions. When there are many semantic errors then the developer may focus on fine-tuning the model with a focus on the meaning of the natural language or choose to flag common semantic errors in the prompt. 

\subsection{Semantic Error Detection and Correction}
\label{sec:SEDAC}
In the following we describe our method for analyzing and auto-correcting errors according
to our error framework. We start with a brief overview of the main ideas and its core
algorithm, \pr{SEDAC} (\emph{Semantic Error Detection And Correction}) shown by the blue box in Figure \ref{fig:LLMtoolpipeline}.

\pr{SEDAC} takes as input a natural language script \id{nl} and the string representation of a logic program
\id{lp}. The \id{nl} is the original problem statement and, in this sense, holds the
``ground truth''. The \id{lp} is meant as a faithful representation of \id{nl} as obtained
by a given LLM.
The purpose of \pr{SEDAC} is to assess the correctness of the \id{lp} wrt.\ the \id{nl} in
terms of 
the error categories defined above. It also carries out fixes for problems spotted along the way.

\pr{SEDAC} first tries to automatically fix syntactic errors. Correct or fixed statements then
proceed to the semantic error detection phase; statements with un-fixable syntax errors are ignored.  We distinguish between \emph{partial} and \emph{full} error
detection (and correction).  These are (potentially incomplete) operational realizations
of the shallow and deep error categories introduced above, respectively.
Partial error detection is concerned with unsuitable formal
representation of adjectives or nouns that can be discovered confidently on linguistic grounds.
Sophisticated tools like spaCy\footnote{\url{https://spacy.io}} can help with this process.
Full error detection is concerned with discovering more speculative logical errors such as wrong
introduction or removal of negation, and reversed implication.
Correspondingly, discovered shallow errors are always corrected without further
validation, discovered deep errors require correctness validation wrt.\ the given
\id{nl}  possibly in conjunction with an external trusted source for domain knowledge.

Technically, \pr{SEDAC} takes the facts and rules $p$ of \id{lp}
and checks them one by one with a logic representation of \id{nl} and computes a status
  \pr{OK}, \pr{NonFixableError} or \pr{FixableError}. The status of $p$ is
  obtained by a soundness check: if \id{nl} entails $p$ in first-order logic then
  \id{p}'s status is \pr{OK}, 
  otherwise a \fn{propose} function is called that returns candidate fixes for
  $p$ which are again checked for soundness. Among all proposed sound
  fixes, if any, some ``best'' fix is noted with $p$ as a \pr{FixableError}.
  A best fix is one that maximizes the number of \id{nl} statements entailed by a
  tentatively fixed \id{lp}.\footnote{It is tempting to instead require ``completeness'', i.e., the
  converse of the sondness entailment. This criterion would be too strong in practice in
  many cases, as the  \id{lp} might lack some formulas but still entail the query.}
  If no sound fix is produced
  then \id{p}'s status is \pr{NonFixableError}.
  Figure~\ref{fig:SEDAC} shows the ``full'' version of \pr{SEDAC}. There is also a ``partial''
  version described below.
\begin{figure}[htpb]
\caption{The \pr{SEDAC} algorithm in pseudocode. 
}
  \centering
\begin{algorithm}{Semantic Error Detection and Correction, \pr{Full-SEDAC}(\id{nl}, \id{lp})}{sedac}

  \textbf{Input:} A PRONTOQA problem \id{nl} and its translation into a logic
  program \id{lp} by an LLM. 

  \textbf{Output:} A status report for every fact and rule of \id{lp}.

  \begin{pseudo}[fullwidth]
    \rlap{$\id{nl_ax} = \{  \fn{nl_to_fof}(s) \mid \text{\nf $s \in \id{nl}$ and $s$ is not a
        query} \}$} & \quad Natural language as FOL\\
    \rlap{$\id{lp_ax} = \{  \fn{lp_to_fof}(r) \mid \text{\nf $r \in \id{lp}$ and $r$ is not a
        query}  \}$} & \quad Logic program as FOL\\
    $\id{lp_ax_status} = \{\}$ & Result status maps for \id{lp} \\
    for $f \in \id{lp_ax}$ & Soundness: check if LP entailed by NL  \\+
    if $\id{nl_ax} \models f$ & Check next fact or rule $f$ \\+
    $\id{lp_ax_status}[f] = \pr{OK}$ & Record OK status of $f$ \\-
    else & Find best modification of $f$, if any \\+ 
    $\id{cand_fixes} = \{ f' \in \id{propose}(f) \mid \id{nl_ax} \models f'\}$ & Get modifications and
    keep sound ones\\
    if $\id{cand_fixes} \== \emptyset$ & No such modifications exist\\+
    $\id{lp_ax_status}[f] = \rlap{\pr{NonFixableSemanticError}}$\\-
    else \\+
    $\id{f_best} = \arg \max_{f' \in \id{cand_fixes}} \fn{score}(f')$ & $\id{f_best}$
    maximizes entailment of \id{nl_ax} \\+
    where $\fn{score}(f') = \rlap{$|\{  g \in \id{nl_ax} \mid (\id{lp_ax}\setminus\{f\}) \cup \{f'\} \models g\}|$}$\\-
    $\id{lp_ax_status}[f] = \rlap{$\pr{FixableSemanticError}(\id{f_best})$}$\\---
    return \id{lp_ax_status}
  \end{pseudo}
\end{algorithm}
  \label{fig:SEDAC}
\end{figure}
\paragraph{The \fn{nl_to_fof} function.} \pr{SEDAC} calls $\fn{nl_to_fof}(s)$ for translating a natural language
sentence $s$ into first-order logic. We implemented $\fn{nl_to_fof}$ as a definite clause
grammar (DCG) in (SWI-)Prolog. The grammar was reverse-engineered from PRONTOQA
examples\footnote{We found this easier than trying to modify the PRONTOQA code for
  emitting first-order logic formulas. It also more useful in view of re-usability to
  domains that are not synthetically made.}; the
syntactic elements nouns, verbs and adjectives we retrieved from the PRONTOQA source
code. The grammar recognizes quantifiers (determiners) like "each", "any", "every", "a" and tolerates
singular/plural formulations. As a side effect, the natural language parser emits first-order logic formulas.
The parser adjusts nouns in plural form to singular form, as unary predicates.
For example, either sentence ``Cats swim.'' and ``Every cat swims'' become $\forall\, x \ \cn{cat}(x)
\rightarrow \cn{swim}(x)$. Adjectives are normalized into nouns, e.g., $\cn{even_number}(x)$ instead of $\cn{even}(x)$.
This way the grammar defines a canonical logical form for PRONTOQA sentences.
This form is taught to and expected from the LLM translation as well.

\paragraph{The \fn{lp_to_fof} function.}  \pr{SEDAC} calls
$\fn{lp_to_fof}(r)$ for translating a string representation \id{lp} of a logic program
suggested by the LLM into FOL. If the program contains symbol errors, natural language errors or
communication errors, an automated fix is attempted by a python script.
The resulting statements are parsed and translated into FOL one by one. Parsing may fail as not all syntax errors
will always be caught. If it succeeds, translation into FOL is merely
syntax rewriting; if it fails then statement is ignored (taken as `true').

\paragraph{The \fn{propose} function.} The \fn{propose} function takes a FOL formula $f$
and returns a possibly empty set of \emph{proposal formulas}. The algorithm is presented
as a set of rewrite rules ``$\Leftrightarrow$'' and derivation rules ``$\Rightarrow$'' in
Figure~\ref{fig:inference-rules}.
\begin{figure}[htpb]
\caption{The rule system of \fn{propose} for fixing shallow semantic errors (above the double
  lines) and deep semantic errors (below the double lines).}
  \centering
  \let\PBS=\\

  \noindent
\begin{tabularx}{\textwidth}{@{}L@{\hspace*{-3.5ex}}C@{\hspace*{-3.5ex}}L@{\hspace*{1ex}}p{7.8em}X@{}}
  \multicolumn{1}{l}{\bf Premise} & \multicolumn{1}{l}{\bf Kind} & \multicolumn{1}{l}{\bf
    Conclusion} & \multicolumn{1}{l}{\bf Condition} & \multicolumn{1}{l}{\bf Example}\\\hline
  \forall\, x\ p(x) \to f & \Leftrightarrow & \forall\,x\ (x=p) \to f & $p$ is proper noun &
  $\forall\, x\ \cn{tom}(x) \to \cn{swims}(x)$ $\Leftrightarrow$\PBS $\forall\,x\ (x=\cn{tom}) \to \cn{swims}(x)$\\\hline
  [\neg]p(\id{ns}) & \Leftrightarrow & \forall\,x\ \id{n}(x) \to [\neg]p(x) & $\id{ns}$ is the plural form of a
  noun $n$ & $\neg\cn{swims}(\cn{cats})$ $\Leftrightarrow$ \PBS $\forall\,x\ \cn{cat}(x) \to \neg \cn{swims}(x)$\\\relax
  [\neg]p(\id{n}) & \Leftrightarrow & \forall\,x\ \id{n}(x) \to [\neg]p(x) & $\id{n}$ is a singular noun  & $\neg\cn{swims}(\cn{cat})$ $\Leftrightarrow$\PBS $\forall\,x\ \cn{cat}(x) \to \neg \cn{swims}(x)$\\\relax
  [\neg]p(\id{a}) & \Leftrightarrow & \forall\,x\ \id{n}(x) \to [\neg]p(x) & $\id{a}$ is an adjective form of a
  noun $n$ & $\cn{floral}(\cn{even})$ $\Leftrightarrow$\PBS $\forall\,x\ \cn{even_number}(x) \to  \cn{floral}(x)$\\\hline
  \forall\,x\ \id{ns}(x) \to f & \Leftrightarrow & \forall\,x\ \id{n}(x) \to f & $\id{ns}$ is
  the plural of noun $n \neq \id{ns}$ & $\forall\,x\ \cn{cats}(x) \to \cn{swims}(x) \Leftrightarrow \forall\,x\ \cn{cat}(x) \to \cn{swims}(x)$\\
  \forall\,x\ f \to [\neg]\id{ns}(x) & \Leftrightarrow & \forall\,x\ f \to [\neg]\id{n}(x) &
  \emph{same} & $\forall\,x\ \cn{cat}(x) \to \cn{swims}(x)$ $\Leftrightarrow$ \PBS $\forall\,x\ \cn{cat}(x) \to \cn{swim}(x)$\\\hline
  [\neg]a(t) & \Leftrightarrow & [\neg]n(t) & $a$ is an adjective form of a noun $n$ & $\cn{even}(\cn{tom})$
  $\Leftrightarrow$ \PBS $\cn{even_number}(\cn{tom})$\\
  \forall\,x\ \id{a}(x) \to f & \Leftrightarrow & \forall\,x\ \id{n}(x) \to f & \emph{same} &
  $\forall\,x\ \cn{even}(x) \to \cn{swim}(x)$ $\Leftrightarrow$ \PBS $\forall\,x\ \cn{even_number}(x) \to \cn{swim}(x)$\\
  \forall\,x\ f \to [\neg]\id{a}(x) & \Leftrightarrow & \forall\,x\ f \to [\neg]\id{n}(x) &
  \emph{same} & $\forall\,x\ \cn{floral}(x) \to \cn{even}(x)$ $\Leftrightarrow$ \PBS $\forall\,x\ \cn{floral}(x) \to \cn{even_number}(x)$\\\hline\hline
  \forall\,x\ f \to p(t) & \Rightarrow & \forall\,x\ f \to \neg p(t) & \emph{none} & $\forall\,x\ \cn{cat}(x) \to \cn{swim}(x)$
  $\Rightarrow$ \PBS $\forall\,x\ \cn{cat}(x) \to \neg \cn{swim}(x)$\\
  \forall\,x\ f \to \neg p(t) & \Rightarrow & \forall\,x\ f \to p(t) & \emph{none} & $\forall\,x\ \cn{cat}(x) \to \neg \cn{swim}(x)$
  $\Rightarrow$ \PBS $\forall\,x\ \cn{cat}(x) \to \cn{swim}(x)$\\
    \forall\,x\ f \to g & \Rightarrow & \forall\,x\ g \to f & \emph{none} & $\forall\,x\ \cn{cat}(x) \to \cn{swim}(x)$
  $\Rightarrow$ \PBS $\forall\,x\ \cn{swim}(x) \to \cn{cat}(x)$\\\hline
\end{tabularx}
  \label{fig:inference-rules}
\end{figure}

Starting from a singleton set comprised of a given formula, the rules are applied
exhaustively, in any order, preferring rewrite rules over derivation rules.
Rewrite rules replace the premise taken from the current set with its conclusion;
derivation rules add to the current set.  The result is the saturated
set without $f$. It is not difficult to see that this procedure always terminates.

Rewrite rules are meant for shallow error correction.  They revolve around normalization
of plural into singular forms, adjectives into nouns, and proper nouns from type positions
(predicates) to individuum positions (terms). The derivation rules for deep error
correction are more of a speculative kind. We use them for replacing the direction of an
implication and complementing literals. These are general rules, not specific to
PRONTOQA, but informed by the kinds of errors we observed LLMs make.

\paragraph{Reasoning Complexity and Partial \pr{SEDAC}.}
In our highly controlled and closed PRONTOQA environment with its simple formula structure, full
error detection poses no problem. The FOL fragment is Bernays-Sch\"onfinkel logic which is
decided by our ATP Beagle \cite{baumgartner-beagle-2015}. Each entailment proof obligation was decided in very short
time(< 1sec). The sets \id{nl_ax} and \id{lp_ax} have at most 20 formulas each for a given
problem. In the worst case, four candidate fixes are proposed per rule or fact, yielding a
maximum of 20 + 4*20 = 100 ATP calls. We investigated 440 problems with Full-SDEDAC which
took 12h. This time could be shortened considerably by avoiding file-based ATP interface
and with a faster ATP.

More realistic settings have open-world character where the problem statement does not
contain full domain information and ``ground truth oracles'' may not be available.  This
let us chose first-order logic semantics for the soundness tests; a closed world
semantics seems too credulous for entailments (let alone having a highly undecidable
entailment problem).  As a
trivial example, a formula with a syntactic error is always dropped and, this way, could
support an unintended entailment with a default negation inference.  While the ``tool''
could, say, employ logic programming for query answering, deep error fixes should be
proposed cautiously and only if deductively valid.

These considerations motivated us to evaluate two versions of \pr{SEDAC}: the full version
defined above, and a \emph{partial} version for shallow error correction.
More precisely, partial-SEDAC differs from \pr{Full-SEDAC} in that
it receives the \id{lp} only (no \id{nl}) and then immediately calls \fn{propose}
restricted to rewriting-rule error correction only. The result of the partial-SEDAC call
is the result of the \fn{propose} call if not empty (i.e.,\ propose was effective),
otherwise it is the given \id{lp}. 
(We do not provide pseudo-code here.)
These two version allowed us to assess the tradeoffs in effectiveness and expressivity. We
report on the results in 
Section~\ref{sec:results} below. 

\paragraph{Example.}
We demonstrate \pr{Partial-SEDAC} and \pr{Full-SEDAC} with a small example that
we compiled from actual PRONTOQA problems and LLM translations. The example consists of the 
sets \id{nl} and \id{lp} shown on the left of the following table, which are converted to
\id{nl_ax} and \id{lp_ax} shown on the right, respectively, in the 
first steps of (Full-)\pr{SEDAC}. Here and below, FOL formulas are written in TPTP FOF
syntax~\cite{sutcliffe-logic-2023}.

\newsavebox{\nlStr}
\begin{lrbox}{\nlStr}
\begin{minipage}[t]{4.5cm}
\begin{minted}{text}
Each integer is not fruity.
Negative numbers are brown.
Wren is an integer.
True or false: Wren is not fruity.
\end{minted}
\end{minipage}
\end{lrbox}

\newsavebox{\lpStr}
\begin{lrbox}{\lpStr}
\begin{minipage}[t]{4.5cm}
\begin{minted}{text}
even(X) :- integer(X), 0 is X mod 2.
integer(X) :- fruity(X).
integer(wren).
integer(X).
brown(negative). 
?- \+ fruity(wren).
\end{minted}
\end{minipage}
\end{lrbox}

\newsavebox{\nlAx}
\begin{lrbox}{\nlAx}
\begin{minipage}[t]{4.5cm}
\begin{minted}{prolog}
! [A] : (integer(A) => ~ fruity(A))
! [A] : (negative_number(A) => brown(A))
integer(wren)
% Query ~ fruity(wren) ignored
\end{minted}
\end{minipage}
\end{lrbox}

\newsavebox{\lpAx}
\begin{lrbox}{\lpAx}
\begin{minipage}[t]{4.5cm}
\begin{minted}{prolog}
% Syntax error line ignored
! [X] : (fruity(X) => integer(X)))
integer(wren) 
! [X] : integer(X)
brown(negative)
% Query ~ fruity(wren) ignored
\end{minted}
\end{minipage}
\end{lrbox}

\begin{tabularx}{1.0\linewidth}[t]{@{}r@{\hspace{2.5em}}X@{\hspace{0.2em}}X}
  & & \id{_ax} \\\hline
\addSpace{\nlStr}{1ex}{2ex}  \id{nl} & \usebox{\nlStr} & \usebox{\nlAx} \\
\addSpace{\lpStr}{1ex}{2ex} \id{lp} & \usebox{\lpStr} & \usebox{\lpAx} \\\hline
\end{tabularx}

 Our parser for the FOL versions of \id{nl} connects
adjectives/noun pairs into single-name predicates, e.g., as in \pl{negative_number(X)}.
Shallow error correction is designed to align logic programs with this convention.
Notice the attempt to bring in ``background knowledge'' \txt{0 is X mod 2} by the LLM
on line 1 of \id{lp_ax} without instructing to doing so; we classify this into the sub-category of Knowledge Error.

\newsavebox{\lpAxPartialFix}
\begin{lrbox}{\lpAxPartialFix}
\begin{minipage}[t]{0.9\linewidth}
\begin{minted}{prolog}
% Syntax error line ignored
! [X] : (fruity(X) => integer(X))
integer(wren)
! [X] : integer(X))
! [I] : (negative_number(I) => brown(I))
% Query ~ fruity(wren) ignored
\end{minted}
\end{minipage}
\end{lrbox}

\newsavebox{\lpAxFullFix}
\begin{lrbox}{\lpAxFullFix}
\begin{minipage}[t]{0.9\linewidth}
\begin{minted}{prolog}
% Syntax error line ignored
! [X] : (fruity(X) => ~ integer(X))
integer(wren)
% ! [X] : integer(X) is NonFixableError
! [I] : (negative_number(I) => brown(I))
% Query ~ fruity(wren) ignored
\end{minted}
\end{minipage}
\end{lrbox}

The FOL resulting from the \pr{SEDAC} runs are as follows:
\begin{center}
\begin{tabularx}{1.0\linewidth}{@{\hspace{0.5em}}rX@{\hspace{2.2em}}X}
\qquad &   $\pr{Partial-SEDAC}(\id{lp})$ & $\pr{Full-SEDAC}(\id{nl}, \id{lp})$ \\\hline
\addSpace{\lpAxPartialFix}{1ex}{2ex}  & \usebox{\lpAxPartialFix} & \usebox{\lpAxFullFix}\\\hline
\end{tabularx}
\end{center}

It is instructive to compare the results of partial and full \pr{SEDAC}.
\pr{Partial-SEDAC}(\id{lp}) differs from \id{lp_ax} only on line 5 by noun and adjective
corrections. \pr{Full-SEDAC}(\id{nl}, \id{lp}) includes this fix as well. In addition, it
fixes the formula $f = \pl{! [X] : fruity(X) => integer(X)}$ on line 2 of \id{pl_ax} by negating its
conclusion. This happens in three steps. First, the
entailment check on line 5 in \pr{Full-SEDAC} finds $\id{nl_ax} \not\models f$.
Then, $\id{propose}(f)$ returns four variants of $f$ but only $f' = \pl{! [X] : fruity(X) => ~
  integer(X)}$ satisfies $\id{nl_ax} \models f'$. Scoring is irrelevant in this case.
The status for $f$, hence, is \pr{FixableError}. As a further difference, the
formula $f = \pl{! [X] : integer(X)}$ on line 4 of \id{lp_ax} has status
\pr{NonFixableError} as $\id{nl_ax} \not\models f$ and no fix is proposed.

Now consider the query \txt{True or false: Wren is not fruity}. The correct answer is
\txt{True} as $\id{nl_ax} \models q$ where $q = \pl{~ fruity(wren)}$. The LLM translation cannot
show that ($\id{lp_ax} \not\models q$), neither can the partial fix ($\pr{Partial-SEDAC}(\id{lp}) \not\models
q$) but the full fix can ($\pr{Full-SEDAC}(\id{nl}, \id{lp}) \models q$).

\section{Results}
\label{sec:results}

Table \ref{tab:Accuracy} shows the overall accuracy of all three models with each experimental condition described in Section \ref{sec:methods}. The results show that the use of the LP system, Fusemate, increased the accuracy of each LLM by between 10\% and 25\% of the possible total. 

\begin{table}[!h]
  \caption{Accuracy for each technique for each model type. Random guessing would be
    expected to achieve an accuracy of $0.5\pm0.05$. The error values
    are half the range across three trials.}
  \centering
  \begin{tabular}{|>{\centering\arraybackslash}m{5cm}|c|c|c|}
    \hline
    Prompt Strategy & GPT3 & GPT4 & Gemini-Pro \\
    \hline
    \multirow{1}{*}{Normal} & $ 0.48\pm0.06 $ & $ 0.83\pm0.12 $ & $ 0.47\pm0.04 $ \\
    \hline
    \multirow{1}{*}{Chain of Thought + one-shot} & $ 0.65\pm0.15 $ & $ 0.94\pm0.04 $ & $ 0.74\pm0.12 $ \\
    \hline
    \multirow{1}{*}{Fusemate} & $ 0.66\pm0.05 $ & $ 0.94\pm0.015 $ & $ 0.57\pm0.03 $ \\
    \hline
    \multirow{1}{*}{Fusemate + one-Shot} & $ 0.76\pm0.06 $ & $ 0.94\pm0.015 $ & $ 0.67\pm0.03 $ \\
    \hline
    Fusemate + one-shot + syntax fix  & $0.83\pm0.06$ & $0.95\pm0.02$ & $0.74\pm0.02$ \\
    \hline
    Fusemate + one-shot + partial fix & $0.87\pm0.05$ & $0.983\pm0.005$ & $0.77\pm0.04$ \\
    \hline
    Fusemate + one-shot + full fix & $0.98\pm0.01$ & $0.995\pm0.005$ & $0.96\pm0.04$ \\
    \hline
  \end{tabular}
  \label{tab:Accuracy}
\end{table}

The \pr{SEDAC} auto-correction successfully reduced errors in all cases. The syntactic fix alone reduced the number of errors of each model by $15-30\%$. The partial and full semantic fixes reduced the number of model errors by $45-72\%$ and $88-92\%$ respectively. In addition to error correction, the \pr{SEDAC} algorithm also classifies the types of errors. 

\begin{table}[!h]
  \caption{Error breakdown for each model type. For each entry, the left value is the average number of each type of error from 100 problems. The values given on the right are half of the range across the three trials for each experimental condition.}
  \centering
  \small
\begin{tabular}{@{}p{1.6cm}p{1.1cm}p{1.1cm}p{1.1cm}p{1.1cm}p{1.1cm}p{1.3cm}p{1.3cm}p{1.2cm}@{}}
\toprule
Techniques & Commun- ication Errors & Symbol Errors & Knowl- edge Errors & Natural Language Errors & Other Syntax Errors & Shallow Semantic Errors & Deep Semantic Errors & Total Instances with Errors \\
\midrule
GPT3  & 0.0±0.0 & 1.3±0.5 & 2.0±1.5 & 3.7±1.5 & 0.3±0.5 & 12.0±1.5 & 32.3±3.0 & 34.3±5.5 \\
\footnotesize{1ShotGPT3} & 0.0±0.0 &3.3±1.5 &0.3±0.5 & 1.7±0.5 & 0.0±0.0 & 10.0±2.5 & 19.3±4.5 & 24.3±4.0 \\
GPT4  & 0.7±0.5 & 1.0±1.0 & 0.0±0.0 & 0.0±0.0 & 0.0±0.0 & 4.7±1 & 1±1 & 6.0±1.5 \\
\footnotesize{1ShotGPT4} & 0.0±0.0 & 0.0±0.0 & 1.0±1.0 & 0.0±0.0 & 0.0±0.0 & 3.7±1.0 & 1.67±0.5 & 6.0±1.5\\
Gemini  & 5.7±1.5 & 4.0±1.5 & 4.3±0.5 & 3.7±2.0 & 1.0±1.0 & 18.7±0.5 & 36.7±2.5 & 43.3±3.0 \\
\footnotesize{1ShotGemini} & 0.7±1.0 & 2.7±3.0 & 5.3±2.5 & 1.0±1.0 & 0.0±0.0 & 16.7±2 & 27.7±0.5 & 32.7±3.0 \\
\bottomrule
\end{tabular}
    
  \label{tab:ErrorTypes}
\end{table}

\normalsize

For each of the Fusemate methods, the types of errors were determined as described in the Sections \ref{sec:errorcat} and \ref{sec:SEDAC}. Table \ref{tab:ErrorTypes} shows the average frequency of each error type across $n=100$ test examples. For each of the three models the most common error type is the Shallow Semantic Errors. Communication Errors, Symbol Errors, Natural Language Errors and Other Syntax Errors were decreased by introducing the example prompt. The one-shot case did not reduce the number of semantic errors for the GPT4 model, however it did reduce semantic errors by approximately 30\% for GPT3 and Gemini. 

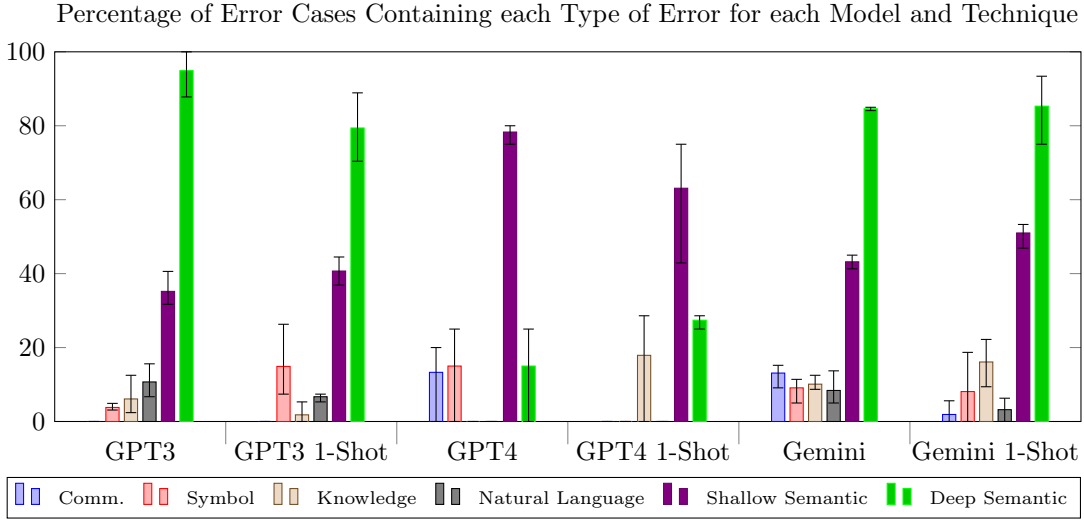
\begin{figure}[!h]
    \caption{A graph of the percentage of error cases that contained each error type for each model. Note that this is an indication of the relative frequency of each error type for a given model and experimental condition. Error bars show the minimum and maximum values across the three trials.}
    \centering
    \hspace*{-1.5ex}
    \begin{tikzpicture}
\begin{axis}[ybar,
        bar width=5pt,
        width=1.05\textwidth,
        title={Percentage of Error Cases Containing each Type of Error for each Model and Technique},
        legend style={at={(1,-0.15)}, anchor=north east, legend columns=-1, font=\scriptsize, /tikz/column sep=5pt},
        ymin=0,
        ymax=100,        
        xtick pos=bottom,
        xtick=data, 
        xticklabels = {
            GPT3,
            GPT3 1-Shot,
            GPT4,
            GPT4 1-Shot,
            Gemini,
            Gemini 1-Shot            
        },
        major x tick style = {opacity=0},
        minor x tick num = 1,
        minor tick length=2ex        
        ]
\addplot+[error bars/.cd, y dir=both, y explicit, error bar style={black}]
    table[x expr=\coordindex, y=Y, y error plus=PosErr, y error minus=NegErr] {\datatableXZero};
\addplot+[error bars/.cd, y dir=both, y explicit, error bar style={black}]
    table[x expr=\coordindex, y=Y, y error plus=PosErr, y error minus=NegErr] {\datatableXOne};
\addplot+[error bars/.cd, y dir=both, y explicit, error bar style={black}]
    table[x expr=\coordindex, y=Y, y error plus=PosErr, y error minus=NegErr] {\datatableXTwo};
\addplot+[error bars/.cd, y dir=both, y explicit, error bar style={black}]
    table[x expr=\coordindex, y=Y, y error plus=PosErr, y error minus=NegErr] {\datatableXThree};
\addplot+[error bars/.cd, y dir=both, y explicit, error bar style={black}]
    table[x expr=\coordindex, y=Y, y error plus=PosErr, y error minus=NegErr] {\datatableXFour};
\addplot+[error bars/.cd, y dir=both, y explicit, error bar style={black}]
    table[x expr=\coordindex, y=Y, y error plus=PosErr, y error minus=NegErr] {\datatableXFive};
\legend{Comm., Symbol, Knowledge, Natural Language, Shallow Semantic,  Deep Semantic}
\end{axis}
\end{tikzpicture}
    \label{fig:ResultsGraph}
\end{figure}

Figure \ref{fig:ResultsGraph} shows the percentage of error cases which contained each error type for each model and technique. This graph shows that the most common errors for GPT3 and Gemini were Deep Semantic Errors, which occur in $75\%$ to $100\%$ of cases. For GPT4 the most common error was Shallow Semantic errors which occurred in approximately $60\%-80\%$ of cases. Note that as the graphed results are normalised, they do not allow for direct comparison of the models' ability to translate the semantic meaning from natural language to logic programs. 

Appendix \ref{sec:appCorMat} contains a correlation matrix for each of the different error types. The matrix shows that most correlations are very weak (magnitude $<0.11$) with only three exceptions. Knowledge Errors show a correlation of $0.23$ with Shallow Semantic Errors, Symbol errors have a $0.31$ correlation with Natural Language Errors and Shallow Semantic Errors anti-correlate ($-0.37$) with Deep Semantic Errors.

\FloatBarrier

Finally, we investigated the effectiveness of our error correction mechanisms. These are `syntactic fixes only', \pr{Partial-SEDAC} and \pr{Full-SEDAC}.
As said earlier, the PRONTOQA problems are agnostic of the reasoning type; with an error-free translation the LP (\id{lp} in Figure~\ref{fig:SEDAC}) is always sufficiently complete in the sense that default reasoning (specifically, default negation) does not enable more conclusions than after reformulation wrt.\ classical first-order logic. 
This is no longer true if the transformation is not correct and the error correction is imperfect. In particular, the corrected \id{lp} may miss relevant rules, which not only removes positive literal conclusions but also adds negative literal conclusions.

For this reason we re-evaluated question answering for different correction scenarios and both open-world and closed-world reasoning. For that, we considered the problems with wrong answers ($n=440$). The results are summarized in Figure~\ref{fig:run-after-fix} which expands on the summarised results in Table \ref{tab:Accuracy}. This table shows that the precision values are systematically higher for the open world semantics compared to the closed world semantics.
\begin{figure}[htpb]
\caption{Re-running divergent problems after syntax only,
  \pr{Partial-SEDAC} and \pr{Full-SEDAC} corrections wrt.\  open-world (classical first-order logic) and closed-world (LP) semantics.}
  \begin{center}
\begin{tabular}{lrrr@{\qquad\qquad}rrr}
& \multicolumn{3}{c}{\bfseries Open-world (FOL)} & \multicolumn{3}{c}{\bfseries Closed-world (LP)} \\
& \multicolumn{1}{l}{Recall} &  \multicolumn{1}{l}{Precision} & \multicolumn{1}{l}{Accuracy} & \multicolumn{1}{l}{Recall} &  \multicolumn{1}{l}{Precision} & \multicolumn{1}{l}{Accuracy} \\\hline
Syntax errors fixed & 0.22 & 0.57 & 0.53 & 0.18 & 0.16 & 0.17\\
\pr{Partial-SEDAC} & 0.38 & 0.72 & 0.63 & 0.35 & 0.32 & 0.34\\
\pr{Full-SEDAC} & 0.80 & 0.98 & 0.89 & 0.85 & 0.81 & 0.83\\
\hline
\end{tabular}
\end{center}
  \label{fig:run-after-fix}

\end{figure}

\vspace{-0.5cm}

\section{Discussion}
The results clearly
show that during the time period of the experiments (December 2023), the accuracy of GPT4 on all experimental conditions was significantly higher than GPT3 and Gemini-Pro which were comparable in their performance. Using an AR tool improved the performance comparable with Chain of Thought techniques and our method has the added bonus of trustworthy explainability; AR tools can produce a proof for any answer they produce. 

For all models, semantic errors were more common than syntax errors. Semantic errors occurred in more than $80\%$ of error cases. 
Therefore the \pr{SEDAC} algorithm showed greater error reduction for semantic errors than syntactic errors. Note that although the  \pr{Full-SEDAC} reduced the total number of errors by $90\%$, most real world scenarios would not have a full semantic fix available. Even in these cases the \pr{SEDAC} algorithm is useful as it allows for classification of errors to rapidly improve prompting. 

As expected, one-shot examples reduced the number of communication errors. Intuitively, providing an example allows the model to better know the required output format. We also expected one-shot to reduce the number of Symbol Errors, this was the case for GPT4, however it made little difference to Gemini and including examples unexpectedly increased the number of Symbol Errors for GPT3. 

Syntax errors occur in a relatively small number of cases compared to semantic errors. This indicates that the capability of state of the art LLMs to produce correct syntax exceeds their ability to express the correct semantics to a tool. This demonstrates the importance of AR tools to enhance the models' reasoning capabilities. We speculate that the `reasoning capacity' of an LLM may be effectively measured by the Chain of Thought accuracy as the corresponding error rate is similar to the total semantic error rate. 

There is currently no prevalent system of classifying types errors in LLM use of tools. There is however one more general error structure which exists in the literature which has some relevance \cite{Xu2023Large}. See Appendix \ref{sec:appErrorClassifcation} for a comparison between this and our error classification.

The results in Figure~\ref{fig:run-after-fix} confirm our expectations that error correction increases recall consistently for open-world and closed-world semantics. Roughly speaking, recall depends mostly on deductive reasoning, which is not as affected by the change of semantics as precision. A high precision value requires a low false positive rate. In our scenario, false positive are often conclusions in the form of negative literals (``True or false: Tom is a not cat'') that become provable by default reasoning when relevant rules are removed by errors. This leads to significantly lower precision than with the open-world semantics. Note that in practice the choice of semantics is mostly likely to be determined by the application domain.

The well defined structure of the natural language in  PRONTOQA allows a DCG to achieve $100\%$ performance. However DCGs are not robust even to small deviations from the assumed structure. Testing LLMs on the PRONTOQA dataset allowed for automated measurement of the frequency and type of LLM errors. We hypothesise that LLMs will be significantly more robust to small changes in wording than DCGs and one area for future work is to test LLM reasoning on unstructured natural language. 

Local LLMs were not used in this study, instead we utilised APIs for pre-trained remote models. Measuring the computational cost is therefore challenging as the structure and number of parameters in each model is not known. Run-time does not provide a reliable measure of computational cost as data transfer and network latency make a varying and significant contributions. A typical response time was 0.5-5 seconds and most responses contained on the order of 100 tokens. One area for future work is to perform similar experiments using local models to accurately determine the computational cost. 

\section{Conclusions}
\label{sec:conclusions}

In this study we have investigated the intersection of Automated Reasoning and Large Language Models in three different ways. Firstly we have explored the capability of LLMs as stand alone reasoning engines. Secondly we have tried coupling LLMs with external Automated Reasoning systems. Thirdly we have implemented automated reasoning technology to debug LLM reasoning. 

We have demonstrated that augmenting an LLM with an AR system improves its reasoning by a similar level to Chain of Thought prompting but with the added bonus of reliable explainability. Furthermore we have introduced the SEDAC algorithm which can act as an auto-correct to reduce LLM errors by at least $15\%$ and up to $90\%$ for problems where a DCG is able to parse the ground truth.

An error classification system was introduced for evaluating interactions between ALMs and their tools. It provides a systematic way to determine the types of errors that LLMs make when interacting with tools. Diagnosing error types provides insight and guidance into which strategies should be implemented to improve model performance. This classification is broad enough that it can be generalised for any external tool while still providing specific information to improve ALM prompts. As the popularity of ALMs rises focus on types of errors gives developers of LLMs a clear direction for improvement. 

One key finding from the paper is that semantic errors are far more common than syntactic errors when LLMs call external tools. This is significant for developers who are interested in deploying LLMs for real-world applications. When prompting their models to use external tools, focus should be placed on enhancing model reasoning and semantics not just syntax. 

This study considers only a restricted domain of steamroller problems which have highly predictable structures. An area for future research is to apply and evaluate these techniques to a broader class of problems or real-world application and 
to determine their computational cost.






\appendix

\section{Comparison with Existing Error Classification Systems}
\label{sec:appErrorClassifcation}
Xu et al.~\cite{Xu2023Large} have two major error categories for determining LLM reasoning capability; evidence selection errors and reasoning process errors. The evidence selection process category is divided into two sub categories which are defined as \cite{Xu2023Large}: 
\begin{itemize}
    \item \textit{Wrong Selection} - `LLMs select the wrong facts or ignore the necessary facts from the beginning of the reasoning.'
    \item \textit{Hallucination} - `LLMs select the evidence which contradicts the given context or cannot be verified from the context.'
\end{itemize}
Note that these categories combined roughly correspond to Knowledge Errors and Deep Semantic Errors. 

Furthermore the reasoning process errors are divided into three sub-categories; no reasoning, perspective mistake and process mistake. In our context the model is not required to reason  per se, instead it is required to translate natural language to a logic program. This best approximates the Shallow Semantic Errors as these clearly indicate a failure in logical reasoning. The communication, symbol and natural language errors have no equivalent error in the system proposed by Xu et al. As the two systems of errors only have rough corresponding categories, any comparison of the frequency error categories should only be a rough approximation. This breakdown would give the results displayed in Table \ref{tab:ErrorCategoryComparison}.

\begin{table}[!h]
  \caption{This table compares the relative frequency of error categories found by this experiment and those reported by Xu et\ al.\ \cite{Xu2023Large}. Note no uncertainty values were reported for the relative frequency of the corresponding error categories. Note that only the GPT3 results were included in this comparison as they most accurate reflection of the models in the review.}
  \centering
  \smallskip
  
{\small
  \begin{tabular}{cc>{\centering\arraybackslash}m{4.2cm}c}
  \hline
    Literature Error & Relative Frequency & \multirow{1}{*}{Corresponding} & Average Relative  \\
    Categories &  & Error Types & Frequency for GPT-3 \\
    \hline
    Hallucination and & \multirow{2}{*}{$ 60.7\%  $} & \multirow{1}{*}{Knowledge Errors and} & \multirow{2}{*}{$ 73\pm15 \%  $} \\
    Wong Selection &  & Deep Semantic Errors &  \\
    \multirow{2}{*}{Perspective Mistake} & \multirow{2}{*}{$ 44.5\% $} & \multirow{2}{*}{Shallow Semantic Errors} & \multirow{2}{*}{$ 52\pm6 \% $} \\
      &  &  &  \\
    \hline
\end{tabular}}
  \label{tab:ErrorCategoryComparison}
\end{table}

Note that the results reported by Xu et\ al.\ would not consider syntactic errors types (except for knowledge and other syntactic errors) as they do not indicate any error in reasoning, only interfacing with an external tool \cite{Xu2023Large}. Their study found that the total number of types of errors per failure was $1.61$; our result for this value is comparable at $1.55 \pm 0.06$.


\section{Example LLM Prompt}
\label{sec:appPrompt}
One of the PRONTOQA steamroller problems reads as follows:\footnote{\ttfamily
  GPT3Fusemate_hops_2.txt, Iteration: 50} 
\begin{quote}
\small{
\itshape Each composite number is not liquid. Every composite number is a fraction. Every
composite number is a number. Negative numbers are not large.
Every fraction is large. Each fraction is a real number. Fractions are integers. Integers are temperate. 
Each number is slow. Each even number is loud. Even numbers are natural numbers. Alex is
an even number. Alex is a composite number.

True or false: Alex is large.}
\end{quote}
The correct answer is ``True''. This can be seen from two modus ponens inferences. Several irrelevant utterances can be ignored.

The following instructions are given to the LLMs:
{\small
\begin{verbatim}
Convert the given English statements into a Prolog program.
Use the format and rules below, including an example for guidance. 
**Format and Rules:**
- Specific statements become facts: 'Whiskers is a cat' -> 'cat(whiskers).'
- General 'is a' statements become rules: 'All cats are birds' -> 'bird(X) :- cat(X).'
- Use '\+' for negations: 'No bird swims' -> '\+swims(X) :- bird(X).'
- Frame queries with '?-': 'Does Whiskers swim?' -> '?- swims(whiskers).'

**Example:**
English: 'All cats are birds. No bird swims. Whiskers is a cat. Does Whiskers swim?'
Problog: cat(whiskers).
bird(X) :- cat(X).
\+swims(X) :- bird(X).
?- swims(whiskers).
Now, convert the following statements into a Prolog program:
Question: 'Each composite number is not liquid. ...'
Problog Program:
\end{verbatim}
}
(The concrete problem is given in full and only listed in abbreviated form above.)

For example, GPT3 answers with the lines:
{\small
\begin{verbatim}
composite(X) :- number(X), \+prime(X).
liquid(X) :- \+composite(X).
fraction(X) :- composite(X).
number(X).
large(X) :- \+negative(X), fraction(X).
real_number(X) :- fraction(X).
integer(X) :- fraction(X).
temperate(X) :- integer(X).
slow(X) :- number(X).
loud(X) :- even(X).
even(X) :- number(X), X mod 2 =:= 0.
natural_number(X) :- even(X).
alex is large. 
\end{verbatim}
}
Notice the question is translated incorrectly in the last line, it should be a prolog
query \verb+?- large(alex).+ Also the rule for \id{even} is unexpected and considered a
syntax error.



\section{Correlation Matrix}
\label{sec:appCorMat}

\begin{figure}[!h]
    \caption{Error Type Correlation Matrix. This shows that there only two significant correlations and one anti-correlation between the types of errors. There is a strong anti-correlation between Shallow Semantic Errors and Deep Semantic Errors, indicating that there are many examples where only one of these two types occurred. There is a correlation between Natural Language Errors and Symbol Errors and also a correlation between Shallow Semantic Errors and Knowledge Errors. All other correlations between errors types are close to 0.}
    \centering
    \includegraphics[width=0.8\textwidth]{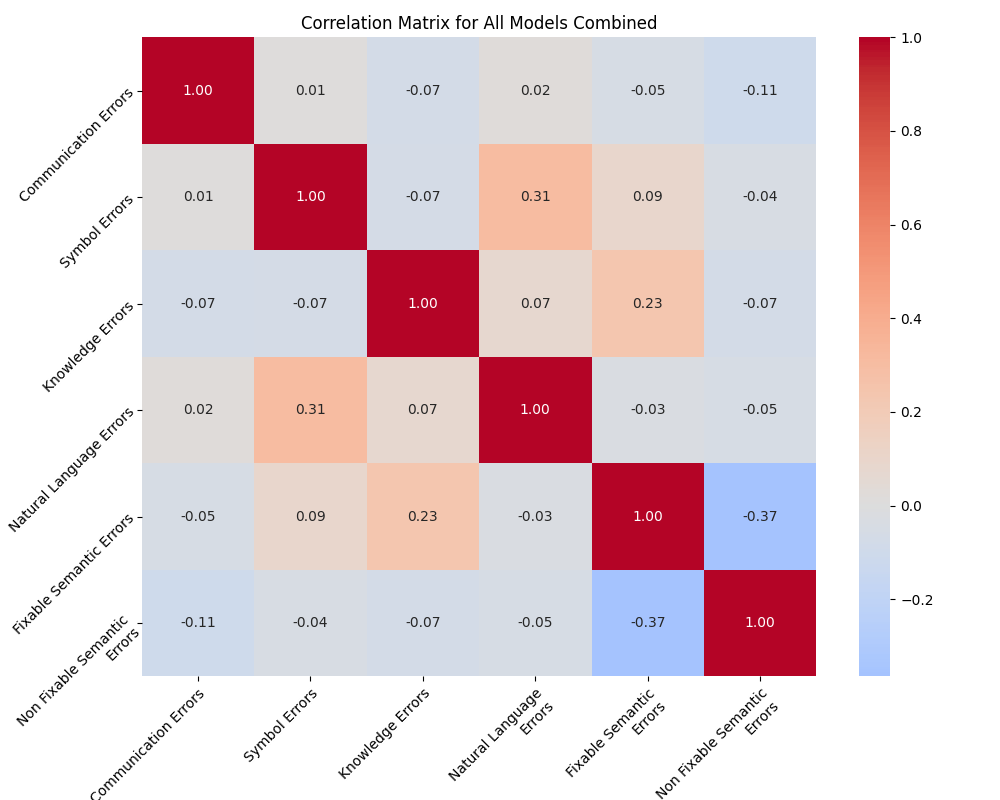}
    \label{fig:CorMat}
\end{figure}

Figure \ref{fig:CorMat} shows the correlation between error types for our experiments. Note that most of the examples that contained errors came from experiments using GPT3 and Gemini, so GPT4 is underrepresented. The correlation between Natural Language Errors and Symbol Errors can be explained by the experimental conditions. In zero-shot examples the model is more likely to make both natural language errors and symbol errors as shown in Figure \ref{fig:ResultsGraph}, while the models make less of these errors in one-shot exmaples. Therefore we would naturally expect to see a correlation between these error types when considering all examples. 

The correlation between Knowledge Errors and
Shallow Semantic has an interesting explanation; it is a feature of the dataset, not the error classification
system.
Knowledge Errors are syntactic errors that cannot be corrected.
Therefore when SEDAC investigates
semantic errors, these lines will always be disregarded.
The results show that for the
remaining lines, higher likelihood that there will be Shallow Semantic Errors.
This can be explained by looking at the most common cause of knowledge errors: inclusion
of mathematical expressions such as \verb|even(X) :- mod2(x)=0|. These problems are also
the most likely problems to mistake adjectives as nouns; for example \verb|prime(X)|
instead of \verb|prime_number(X)| which can also fixed by partial \pr{SEDAC}.

\end{document}